\documentclass[10pt,journal,compsoc]{IEEEtran}

\ifCLASSOPTIONcompsoc
  \usepackage[nocompress]{cite}
\else
  \usepackage{cite}
\fi

\ifCLASSINFOpdf
   \usepackage[pdftex]{graphicx}
\else
   \usepackage[dvips]{graphicx}
\fi

\usepackage{amsmath}

\usepackage{caption}
\usepackage{epsfig}
\usepackage{amssymb}
\usepackage{booktabs}
\usepackage{tabularx}
\usepackage{diagbox}
\usepackage{colortbl}
\usepackage[percent]{overpic}
\usepackage{pstricks}
\usepackage{pst-node}
\usepackage{capt-of,etoolbox}
\usepackage{comment}
\usepackage{multirow}
\usepackage{todonotes}

\begin{document}
\title{A3GC-IP: Attention-Oriented Adjacency Adaptive Recurrent Graph Convolutions for Human Pose Estimation from Sparse Inertial Measurements}

\author{Patrik~Puchert,~\IEEEmembership{Member,~IEEE,}
        and Timo~Ropinski,~\IEEEmembership{Fellow,~IEEE}%
        \setcounter{figure}{0}
\begin{center}
    \centering
    \includegraphics[width=.92\linewidth]{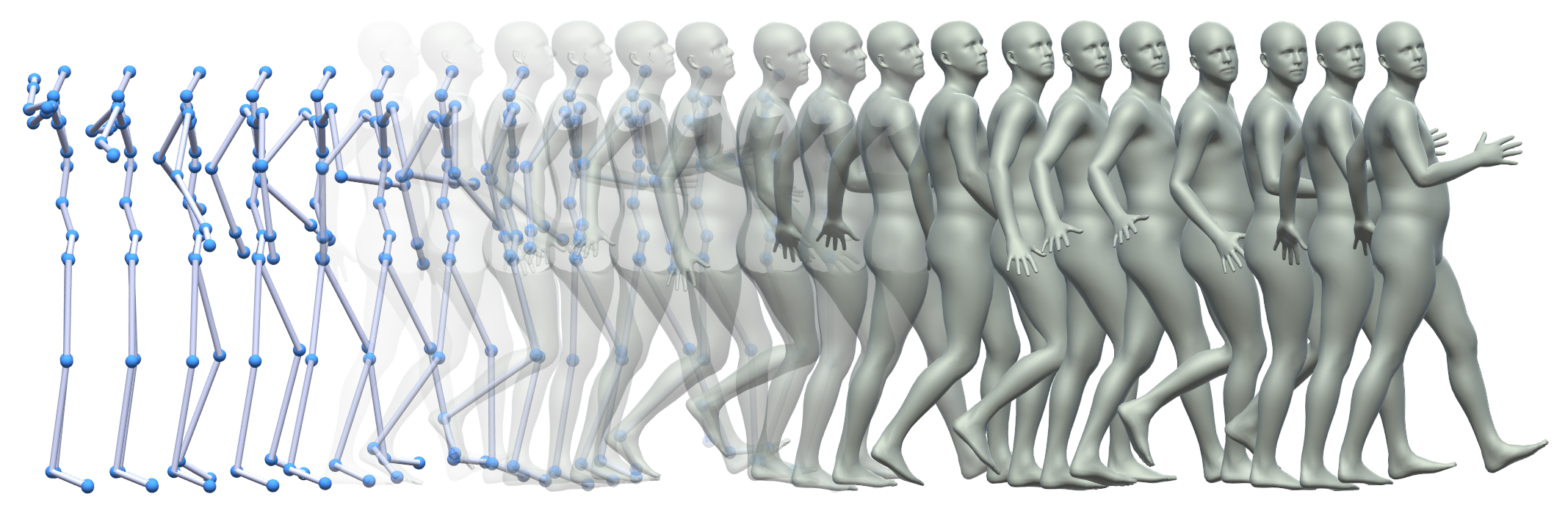}
    \captionof{figure}{We present a new deep recurrent graph network approach to estimate human poses from 6 inertial measurement units (IMUs). Our approach exploits attention-oriented adjacency adaptive graph convolutional long short-term memory cells, to obtain the poses from the normalized IMU data projected onto the skeletal graph. Thus, we increase accuracy on both positional and angular error and outperform the state-of-the-art methods on all evaluated datasets.}
    \label{fig:teaser}
\end{center}%
\IEEEcompsocitemizethanks{\IEEEcompsocthanksitem P. Puchert and T.Ropinski are with the Institute of Media Informatics, Ulm University, Germany.\protect\\
E-mail: patrik.puchert@uni-ulm.de
%E-mail: anonymous@research-institute
\IEEEcompsocthanksitem T. Ropinski is with Institute of Media Informatics, Ulm University, Germany\\
         and Department of Science and Technology, Link\"oping University, Sweden.}% <-this % stops an unwanted space
%\thanks{Manuscript received December 09, 2022; revised TBA.}}
\thanks{}}


\markboth{}%
{Puchert and Ropinski: A3GC-IP - Attention-Oriented Adjacency Adaptive Recurrent Graph Convolutions for HPE from Sparse IMUs}
\IEEEtitleabstractindextext{%
\begin{abstract}
\noindent Conventional methods for human pose estimation either require a high degree of instrumentation, by relying on many inertial measurement units (IMUs), or constraint the recording space, by relying on extrinsic cameras. These deficits are tackled through the approach of human pose estimation from sparse IMU data. We define attention-oriented adjacency adaptive graph convolutional long-short term memory networks (A3GC-LSTM), to tackle human pose estimation based on six IMUs, through incorporating the human body graph structure directly into the network. The A3GC-LSTM combines both spatial and temporal dependency in a single network operation, more memory efficiently than previous approaches. The recurrent graph learning on arbitrarily long sequences is made possible by equipping graph convolutions with adjacency adaptivity, which eliminates the problem of information loss in deep or recurrent graph networks, while it also allows for learning unknown dependencies between the human body joints. To further boost accuracy, a spatial attention formalism is incorporated into the recurrent LSTM cell.
With our presented approach, we are able to utilize the inherent graph nature of the human body, and thus can outperform the state of the art for human pose estimation from sparse IMU data.
\end{abstract}

}

\maketitle

\IEEEdisplaynontitleabstractindextext

% \fi
\IEEEpeerreviewmaketitle

\IEEEraisesectionheading{\section{Introduction}}
\IEEEPARstart{A}{correct} estimation of human poses is important in many applications. These range from various applications in virtual and augmented reality~\cite{10.1007/978-3-642-17688-3_31,rthpdatftrivr} to medical applications, such as gait analysis~\cite{9220146}, patient monitoring ~\cite{achilles2016patient} or human activity recognition ~\cite{AGAHIAN2020196}. 
Unfortunately, today's state of the art (SOTA) methods for human pose estimation either only work in constrained environments, or are very intrusive~\cite{DBLP:journals/corr/abs-2006-01423,zhao2019semantic,zheng20213d}.
These constraints make them impractical for outdoor applications, indoor scenarios spanning multiple rooms or suffering from occlusions~\cite{cheng2019occlusion}.  Measuring the human body pose with body-worn IMUs can solve these deficits~\cite{klenk2019change}. To aid user acceptance and usability, the number of body mounted sensors must be minimal, resulting in sparse inertial measurements.
In this paper, we propose a novel approach for human pose estimation based on a set of 6 IMUs. While this scenario has been tackled by others before~\cite{von2017sparse,huang2018deep,TransPoseSIGGRAPH2021,PIPCVPR2022}, we are the first to enhance pose estimation accuracy by incorporating the structure of the human body through deep graph learning, instead of predicting the pose from a flat array of input data. 
While using graph structures in deep neural networks is a well studied field~\cite{wu2020comprehensive}, the usage of standard graph convolutions in recurrent architectures poses the same over-smoothing problem as in other deep graph architectures~\cite{li2018deeper}. 
To address this challenge, we propose the usage of adjacency adaptive graph convolutions (A2GC) directly inside the recurrent cell. Doing so requires significantly less memory during training, as compared to approaches exploiting graph convolutions with learnable adjacency matrices in different fields~\cite{Li_2020_CVPR}. 
We further equip this new type of LSTM cell with an attention formalism, defining our attention-oriented adjacency adaptive graph convolutional LSTM cells (A3GC-LSTM), with which we are able to outperform the SOTA on sparse IMU driven human pose estimation.
\noindent To further improve pose estimation, we show how A3GC-LSTMs benefit from respecting the bilateral body symmetry, by utilizing contralateral data augmentation. 
This augments the available training data by mirroring all movements, which increases the range of possible motions in the training data and for instance removes any bias of left- or right-handedness in the data.\\
\noindent Thus, within this paper we propose the first graph convolution approach to solve sparse IMU-based pose estimation, and we make the following technical contributions in this context:

\begin{itemize}
    \item We introduce the A3GC-LSTM cell as a memory efficient recurrent graph LSTM formulation incorporating learnable adjacency matrices, to address the over-smoothing problem of recurrent graph convolutional networks.
    
    \item We show how the A3GC-LSTM benefits from utilizing an attention formalism.
    
    \item Using our proposed A3GC-LSTM model we outperform the SOTA for sparse IMU-based human pose estimation. 
    %\item We show how contralateral data augmentation, can enhance the training data by considering symmetry observations.
\end{itemize}

%\noindent By combining these contributions into a single learning approach, we are able to outperform the current SOTA of IMU-based pose estimations performed on sparse inertial measurement data. 
\noindent While we make these technical contributions specifically for IMU-based human pose estimation, we would like to emphasize that they are not restricted to this task and could likely be applied to other pose estimation scenarios without IMUs, as well as for other body types or recurrent tasks employing graphs with constant connectivity.
%as long as the posed model utilizes a rig with constant structure.

\section{Related Work}
Here we briefly review the relevant literature on human pose estimation using cameras and IMUs, as well as relevant work on graph learning and attention.\\

\noindent\textbf{IMU-based pose estimation.} 
Human pose estimation can be tackled by various approaches. 
These range from the setup of multiple calibrated cameras and body worn markers~\cite{canton2010marker} over approaches using RGBD cameras~\cite{zimmermann20183d}, or ultrasonic technology~\cite{laurijssen2016ultrasonic}, to such based on monocular RGB input~\cite{Rhodin_2018_CVPR,Sharma_2019_ICCV,Xu_2020_CVPR}. 
These approaches however all have in common, that they can only operate in a constrained volume given by the field of view of the involved sensors. 
%A system relying on electromagnetic field sensing was proposed in which the receiver component is also body-worn~\cite{Kaufmann_2021_ICCV}. 
%While this approach shares similarities to IMU-based approaches regarding the applicability and intrusiveness, it is also susceptible to disturbances from the surroundings of the recorded subject, while also enhancing the complexity of the body worn setup. 
%Traditional motion Capture technologies constrain the recording volume by relying on extrinsic sensors like cameras~\cite{canton2010marker,zimmermann20183d,laurijssen2016ultrasonic,Rhodin_2018_CVPR,Sharma_2019_ICCV,Xu_2020_CVPR}.
These problems are lifted by completely body-worn systems. 
While also other systems have been proposed~\cite{Kaufmann_2021_ICCV}, IMU based systems have the benefit of being light weight, small and widely available.
Today, such systems are commercially available, whereby IMU measurements from different units are fused to reconstruct a subject's pose~\cite{roetenberg2009xsens}. 
While they yield good results, their large amount of required IMUs, e.g., 17 IMUs in Xsens' current MVN setup for full pose estimation\cite{schepers2018xsens}, can be considered intrusive, require long setup times, and provoke sensor placement errors.
To tackle these shortcomings, several methods have been proposed to reduce the amount of necessary sensors.
Many such approaches are limited in application by matching query samples to prerecorded databases~\cite{slyper2008action,tautges2011motion} or training a model for each activity of interest~\cite{schwarz2009discriminative}.
%To tackle these shortcomings, sparse accelerometer data has been used for motion reconstruction~\cite{slyper2008action,tautges2011motion}. 
%However, these methods work by matching measured input to samples from a prerecorded database, and thus suffer from a limited generalizability. 
%With a similar drawback Schwarz et al. have proposed a learning-based method using Gaussian processes to regress the full body pose from sparse IMU input~\cite{schwarz2009discriminative}.
%As their model is trained individually for each activity of interest performed by an actor, the approach's generalizability is severely limited. 
Von Marcard et al. instead use a generative approach to estimate 3D poses from only 6 IMUs with their sparse inertial poser (SIP)~\cite{von2017sparse}. 
To do so, they equip the Skinned Multi-Person Linear (SMPL) body model~\cite{loper2015smpl} with synthetic IMUs, and solve for the SMPL pose that best matches the measured IMU data. As this has to be done at query time, it is a computationally expensive procedure. 
With the deep inertial poser (DIP) Huang et al. have enhanced the accuracy of pose estimation based on 6 IMUs using a deep learning approach~\cite{huang2018deep}. 
They built a bidirectional LSTM network~\cite{schuster1997bidirectional} to map the flattened input of 5 IMUs, normalized by a 6th IMU on the pelvis, to the target pose in the SMPL body model.
%Butt et al. have build upon this work to predict reliable confidence intervals for human pose estimation~\cite{9363874}. 
In contrast to the deep learning approaches the use of shallow fully connected networks has been proposed~\cite{s19173716}.
Yi et al. proposed an improvement for the deep architectures by predicting intermediate representations, along with a network branch for global position estimation~\cite{TransPoseSIGGRAPH2021}. 
With their Transpose network they effectively apply the network of DIP on three subsequent steps, where they first predict the position of leaf joints from the IMU data, then the position of all joints from the former joints as well as the IMU data, and finally the target pose in the SMPL model using the IMU data and the position of all joints as input. 
Recently, they further improved the accuracy by introducing a physics-based in the physical inertial poser (PIP)~\cite{PIPCVPR2022}. PIP combines the Transpose network with a non-learnable physically driven module to postprocess the model predictions.
Sparse IMU input has also been used in combination with RGB cameras~\cite{von2016human,pons2010multisensor,pons2011outdoor,malleson2017real} as well as RGBD cameras~\cite{helten2013real}. While these approaches show increased accuracy over camera-only methods, they still suffer from the constraints which come with the use of cameras. 
Thus, while traditional work requires an environment constrained by at least one camera, or is limited by existing databases, DIP~\cite{huang2018deep} and Transpose~\cite{TransPoseSIGGRAPH2021} take the next step, by incorporating modern learning approaches. 
While they can be considered SOTA with respect to accuracy, they do not consider the graph nature of the human body as they operate on flattened data. \\

\noindent\textbf{Graph learning.} 
Graph learning has proven beneficial in many disciplines~\cite{wu2020comprehensive}. 
Graph learning has been used for human body related tasks, by mapping a 2D pose, predicted from static RGB images, to a 3D pose~\cite{zhao2019semantic,zou2021modulated}, eliminating the need to cope with time dependencies. 
Several approaches have been proposed which work with spatio-temporal data by separating the spatial and temporal learning steps.
These separations are either a consecutive application of graph convolutions (GCN) and recurrent layers~\cite{zhao2019t,nicolicioiu2019recurrent}, or of GCNs in space and convolutions in time domain consecutively~\cite{yan2018spatial,shi2019two}. 
Some of these methods modulate the fixed adjacency of the underlying graph by incorporating learnable adjacency components.
The combination of spatial and temporal dependencies has been proposed for different tasks. 
Bai et al. build the adjacency matrix out of a learnable embedding of all graph nodes and use them in gated recurrent units (GRUs)~\cite{bai2020adaptive} for traffic forecasting. 
While such an embedding has benefits for large graphs, it is not beneficial for small graphs as encountered in our tasks.
Li et al. have proposed graph-based GRU cells (G-GRU) for human motion prediction, in which they also employ an adaptive adjacency matrix~\cite{Li_2020_CVPR}. To employ this adaptivity, they modulate the fixed adjacency matrix with a multiplicative and an additive weight matrix before using it in a GCN to modify the cells hidden input state. 
While we consider it counter-intuitive to just add a graph computation to the network while keeping the linear computation, it also requires a significantly larger memory footprint than our approach, as we will detail in section \ref{param_comp_section}. 
LSTMs differ from GRUs by having a better deep context understanding~\cite{gruber2020gru}, which makes them better suited to tasks such as human pose estimation, where input sequences can consist of data recorded with 60Hz, while the context of movements can span from only a few frames to many seconds.
While the over-smoothing problem of deep GCN networks~\cite{li2018deeper} has been tackled in the spatial domain~\cite{pmlr-v119-chen20v} by introducing initial residual connections and identity mapping to the GCN, this is not applicable to recurrent architectures. \\

\noindent\textbf{Attention.} 
While the usage of attention has originally gained popularity for natural language processing\cite{vaswani2017attention, chorowski2015attention}, it has since been applied to many different fields\cite{ramachandran2019stand, jaegle2021perceiver}.
Si et al. use a combination of LSTMs and GCNs with fixed adjacency together with an attention mechanism for human activity recognition~\cite{si2019attention}. 
While they obtain good results, their model was defined for short sequences (up to 100 frames) with densely annotated graphs as input, and does not scale well to our problem with indefinitely long sequences and sparse inputs. Thus we combine this spatial attention approach with our A2GC-LSTMs.

\section{The Method}
\begin{figure}[t]
    \center
    \includegraphics[trim=0 1 0 0, clip, width=0.62\linewidth, keepaspectratio]{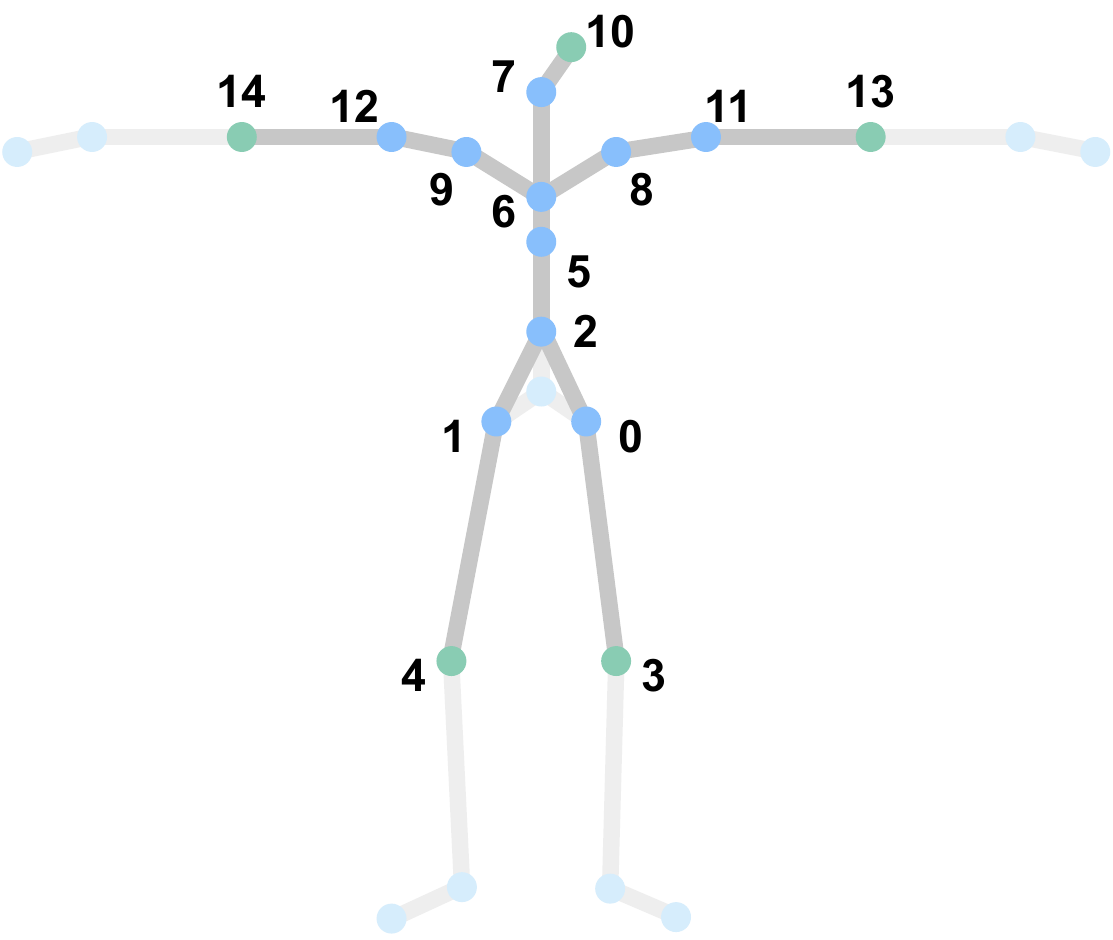}
    \caption{\label{joint_numbers}We learn towards the SMPL skeleton (\emph{light gray}), whereby we limit the joint connectivity (\emph{dark gray}) to match the influence area of the 5 normalized \emph{IMUs}, placed at the green joints.}
\end{figure}
\begin{figure*}[th!]
    \center
    \includegraphics[trim=00 00 00 00, clip, width=\linewidth, keepaspectratio]{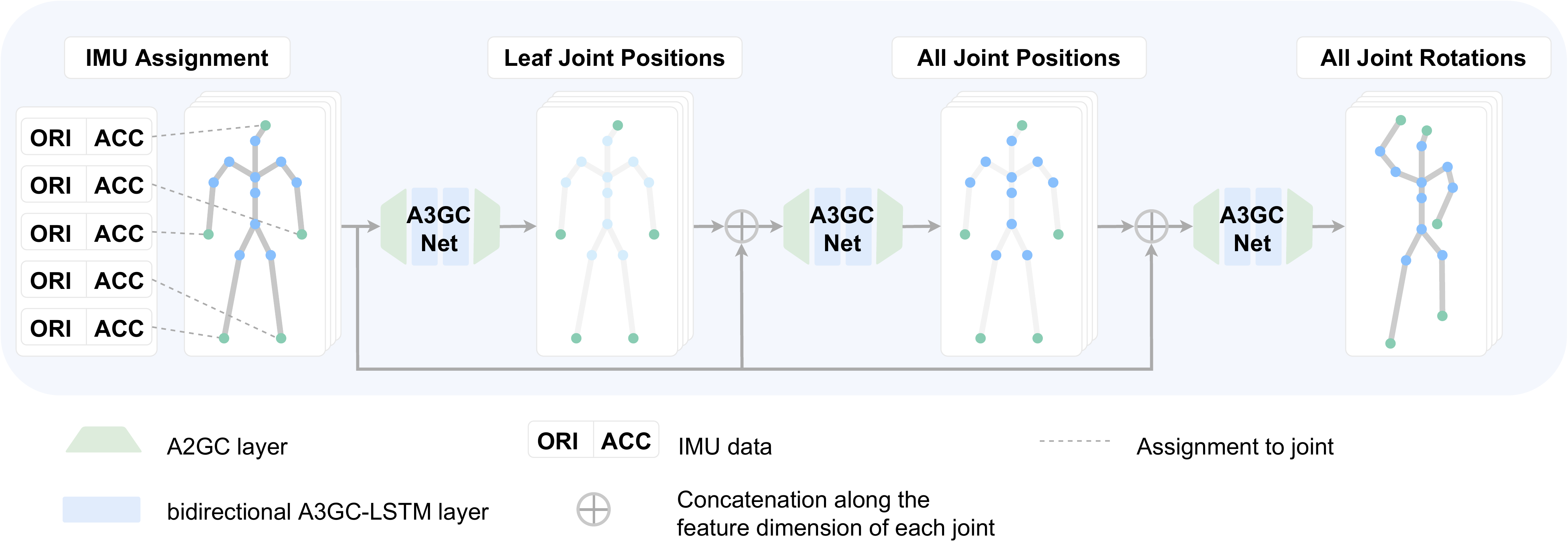}
    \caption{\label{model_figure} Layout of our proposed A3GC-IP network (\emph{top}). The human pose is predicted by A3GC networks in three steps. The first network predicts the positions of the leaf joints, the  second the positions of all joints and the last network predicts the rotations of all joints, i.e. human pose. $[ORI,ACC]$ are the concatenated orientation and acceleration values from the IMUs. On \emph{bottom} we detail the layout of the A3GC net. Each network consists of an A2GC Input layer, two bidirectional A3GC-LSTM layers and one A2GC output layer.}
\end{figure*}
In this section, we detail the technical concepts behind our proposed graph convolution approach. The input of our model is the data of 5 IMUs placed on the SMPL skeletal graph, whereby their input is transformed into a body-centric system as per Huang et al. by a 6th IMU on the pelvis~\cite{huang2018deep}. 
We also train towards the SMPL body model~\cite{loper2015smpl} as target, in order to obtain realistic poses, and allow for comparison to the SOTA. 
To incorporate body topology, we transform the IMU input to a graph structure by automatically placing the sensors on a human skeletal graph at the corresponding nodes. 
For this graph we chose the SMPL skeletal model, whereby we focus on only 15 core joints out of the 24 joints of the model~\cite{loper2015smpl}, i.e., without the outer extremities of hands, feet and the root joint (see Fig.~\ref{joint_numbers}). 
We chose this focus due to the fact, that the information sparse IMUs can give for the former is naturally limited, while the root joint at the pelvis is fixed by definition through the data transformation with respect to it. 
The nodes containing no IMU measurements are initialized with zeros in the input graph. Thus, we operate on an input graph of dimensionality $N\times F_{in}$, where $N=15$ and $F_{in}=12$ are the input features given by the elements of a $3x3$ rotation matrix and a three dimensional acceleration vector. The training target is the same graph with the SMPL pose parameters $\theta$ on the corresponding nodes.\\
%, which define the pose of the SMPL model. 

\noindent\textbf{Adjacency adaptive graph convolution.} 
The core of our model is the combination of LSTM cells and GCNs in a bidirectional recurrent layer. 
As the inclusion of standard GCNs in recurrent applications suffers from over-smoothing problems as described by Li et al.~\cite{li2018deeper}, we introduce adjacency adaptive graph convolution (A2GC) as a remedy. 
The definition of A2GC follows the notation of the commonly used approximation of the graph convolution as proposed by Kipf and Welling~\cite{kipf2016semi}, with the propagation rule:
\begin{equation}
\label{gcn_eq}
    \mathbf{Z} = \tilde{\mathbf{A}}\mathbf{X}\mathbf{W} + \mathbf{b},
\end{equation}
\noindent where $\mathbf{X}$ is the input and $\mathbf{W}$ and $\mathbf{b}$ are the trainable weights and biases. 
In standard GCNs $\tilde{\mathbf{A}} = \mathbf{D}^{-\frac{1}{2}}(\mathbf{A}+\mathbf{I_N})\mathbf{D}^{\frac{1}{2}}$ is the constant symmetric normalization of the adjacency matrix $\mathbf{A}$ with added self-connections $\mathbf{I_N}$ using the diagonal node degree matrix $\mathbf{D}$ of $\mathbf{A}$.
To now employ adjacency adaptivity, we instead make $\tilde{\mathbf{A}}$ a learnable matrix, initialized by the normalized complemented distance on the graph:
\begin{equation}
    \label{ncd_eq}
    \tilde{\mathbf{A}}_{ij}^{init} = 1 - \frac{d(n_i,n_j)}{\sum_j d(n_i,n_j)},
\end{equation}
\noindent where $d(n_i,n_j)$ is the Euclidean distance between node $i$ and node $j$ on the graph. We will show empirically that the adaptivity of adjacency lifts the problem of over-smoothed results in an automated manner (Sec. \ref{Ablation_section}). 
In addition to the necessity of adjacency adaptivity in our application we assume another benefit of the learnable adjacency, which is the hardly factorizable dependency of all joints to each other.
The problem in factorization of all joint dependencies comes from the fact, that biological joints are not only actuators driven by many muscles, but even in a free state underlie many factors of dampening~\cite{minetti2020frictional,leardini2014biomechanics}. 
Thus the free joint, i.e. one not actively driven by muscular movement, is affected by all other joints both through the connection along a line of damped oscillators, as well as by inertia, as the skeleton is neither completely stiff.
While it is true that all effects will always also effect the neighbouring joints, it can be better for the model to learn dependencies to the other joints as well, as this will give the context to small changes in acceleration or rotation.
Using this definition of the A2GC operation inside LSTM cells (A2GC-LSTM) thus not only lifts the over-smoothing problem, but also allows for learning of unknown joint dependencies.
The definition of the A2GC-LSTM cell follows on a coarse level the definition of the conventional LSTM cell~\cite{hochreiter1997long,gers1999learning}, whereby in the A2GC-LSTM cell we replace every learnable network operation with A2GCs:

\begin{equation}
    \label{A2GC-lstm_eq}
    \begin{aligned}
        &\mathbf{X}_i         &&= \sigma(\tilde{\mathbf{A}}_i\mathbf{X}\mathbf{W}_i + \mathbf{b}_i)\\
        &\mathbf{X}_f         &&= \sigma(\tilde{\mathbf{A}}_f\mathbf{X}\mathbf{W}_f + \mathbf{b}_f)\\
        &\mathbf{X}_c         &&= tanh(\tilde{\mathbf{A}}_c\mathbf{X}\mathbf{W}_c + \mathbf{b}_c)\\
        &\mathbf{X}_o         &&= \sigma(\tilde{\mathbf{A}}_o\mathbf{X}\mathbf{W}_o + \mathbf{b}_o),
    \end{aligned}
\end{equation}

\noindent where $\mathbf{X}$ is the concatenation of the current input and the last hidden state $\mathbf{H}_{t-1}$, to both of which dropout is applied, and the index $t$ denotes the timestep. 
$\mathbf{\tilde{A}}$, $\mathbf{W}$ and $\mathbf{b}$ are the adjacency and weight matrices and biases of the A2GC, and $\sigma$ is the sigmoid activation function. The gates $X_{\{i,f,c,o\}}$ are then processed with the common LSTM scheme~\cite{hochreiter1997long}:

\begin{equation}
    \label{lstm_eq}
    \begin{aligned}
        &\mathbf{C}_t         &&= \mathbf{C}_{t-1}\odot\mathbf{X}_i + \mathbf{X}_f\odot\mathbf{X}_c\\
        &\mathbf{H}_t         &&= \sigma_{out}(\mathbf{C}_t)\odot\mathbf{X}_o\\
        &\mathbf{O}_t         &&= \sigma_{out}(\mathbf{H}_t),
    \end{aligned}
\end{equation}

\noindent where $\mathbf{O}$ is the cells output and $\mathbf{C}$ is the cells carry respectively. $\sigma_{out}$ is the output activation function, which we realize as a $tanh$-function. 
Thus, we are able to combine the spatial and temporal learning part in a single recurrent cell, and are able to learn all dependencies of a current state in a single computation. 
%This is desirable as neither spatial, nor temporal dependency can be viewed on its own, since the human body is a dynamic construct moving trough time and has to be viewed as such. 
This is desirable as the spatial context needs to be accessible for every joint to define the pose. Furthermore, since the single pose is only a discrete timestep in a continuous movement, the temporal dependency must be accessible to consider the pose in a coherent context of an animation. 
As such, the pose of a raised arm could inherit from a raising arm movement or of relaxing an arm from a raised pose, in which the sensor rotation values will be identical while the temporal information puts the sparse acceleration data in a temporal context better suited for robust estimation.\\

\noindent\textbf{Attention-oriented A2GC.}
\begin{figure}[t!]
    \center
    \includegraphics[trim=600 00 600 00, clip, width=\linewidth, keepaspectratio]{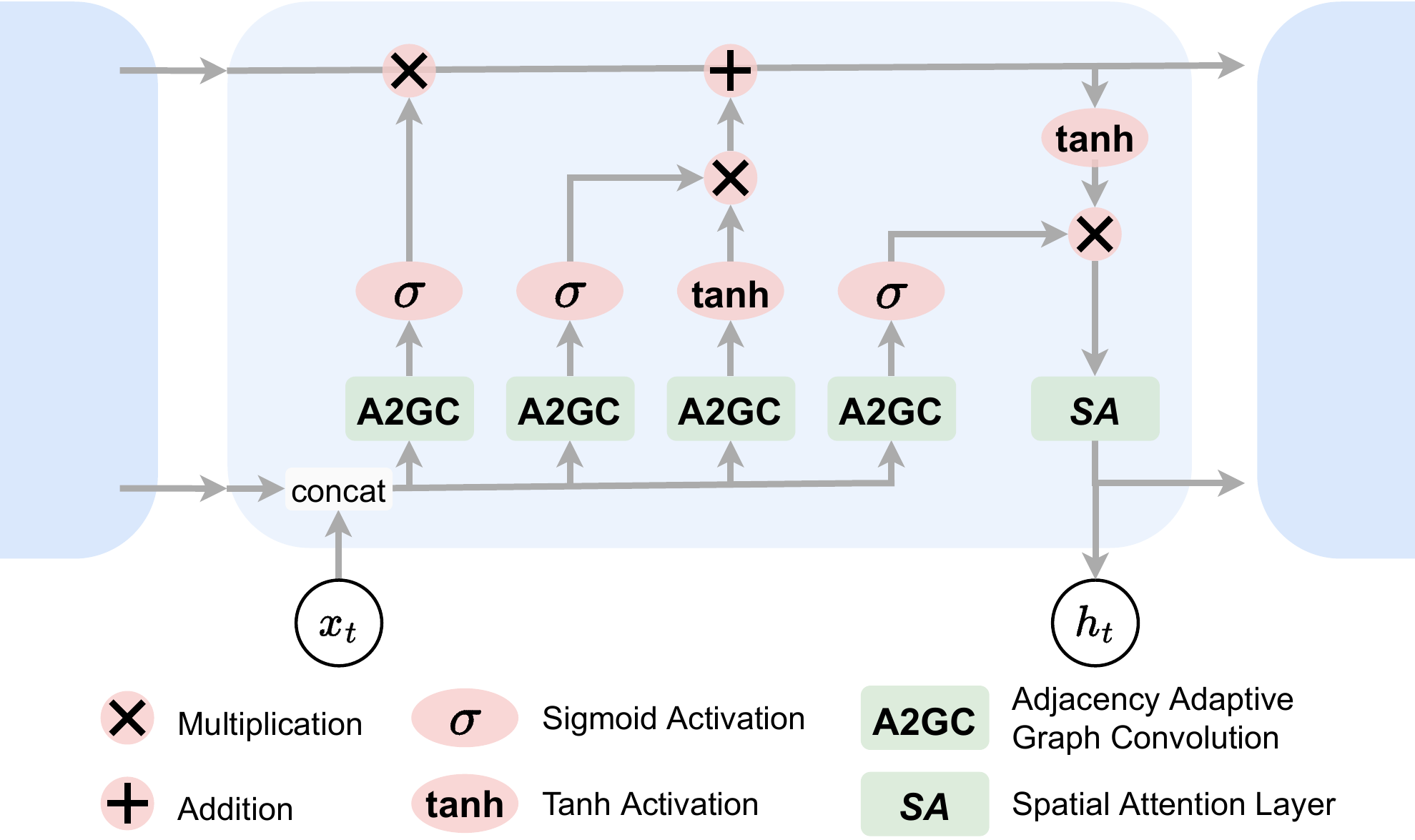}
    \caption{\label{cell_figure} Structure of the A3GC-LSTM cell. The linear operations of the common LSTM cell are replaced with \emph{A2GC} operations and the hidden state is passed through a spatial attention operation (\emph{SA}).}
\end{figure}
\begin{table*}[ht]
    \caption{\label{main_table} Evaluation of the proposed A3GC-IP model compared to DIP, Transpose and G-GRU on DIP-IMU~\cite{huang2018deep} and Total Capture~\cite{Trumble:BMVC:2017}. For DIP we list the result of the method as proposed by the authors as well as a version adopting global target rotations. We report the mean global angular error over the shoulder and hip joints (DIP Err), as well as the mean global angular error, the mean position error and the mean jerk error averaged over all 15 joints.}
    \centering
    \setlength{\tabcolsep}{3.4pt} % Default value: 6pt
    \renewcommand{\arraystretch}{1} % Default value: 1
    \footnotesize
    \begin{tabular}{l cccc c cccc}
        \toprule
        &\multicolumn{4}{c}{\textbf{DIP-IMU}}
        &&\multicolumn{4}{c}{\textbf{Total Capture}}\\
        \cmidrule{2-5}
        \cmidrule{7-10}
                        & \textbf{DIP Err} $[deg]$
                        & \textbf{Ang Err} $[deg]$
                        & \textbf{Pos Err} $[cm]$
                        & \textbf{Jerk Err} $[\frac{km}{s^3}]$
                        && \textbf{DIP Err} $[deg]$
                        & \textbf{Ang Err} $[deg]$
                        & \textbf{Pos Err} $[cm]$
                        & \textbf{Jerk Err} $[\frac{km}{s^3}]$\\
        \midrule
          DIP~\cite{huang2018deep} & $16.98 (\pm 8.94)$ & $13.58 (\pm 7.54)$ & $7.05 (\pm 3.87)$ & $2.32 (\pm 3.36)$ && $\mathbf{16.36 (\pm 9.69)}$ & $14.51 (\pm 7.67)$ & $7.90 (\pm 4.56)$ & $2.41 (\pm 3.15)$\\

          DIP (global)  & $14.03 (\pm 7.19)$ & $7.94 (\pm 4.28)$ & $5.92 (\pm 3.16)$ & $2.69 (\pm 3.85)$ && $25.62 (\pm 8.83)$ & $14.58 (\pm 5.90)$ & $8.66 (\pm 4.27)$ & $2.43 (\pm 3.86)$\\
          
          Transpose~\cite{TransPoseSIGGRAPH2021} & $14.02 (\pm 7.10)$ & $7.46 (\pm 4.04)$ & $5.54 (\pm 2.94)$ & $1.90 (\pm 3.11)$ && $26.87 (\pm 9.08)$ & $15.08 (\pm 6.30)$ & $8.27 (\pm 4.37)$ & $0.70 (\pm 1.30)$\\
          
          G-GRU~\cite{Li_2020_CVPR} & $14.40 (\pm 7.26)$ & $7.80 (\pm 4.05)$ & $6.27 (\pm 3.13)$ & $2.05 (\pm 3.28)$ && $26.07 (\pm 8.05)$ & $14.48 (\pm 5.71)$ & $8.16 (\pm 3.89)$ & $0.99 (\pm 1.89)$\\
          
          A3GC-IP & $\mathbf{13.57 (\pm 6.76)}$ & $\mathbf{7.18 (\pm 3.72)}$ & $\mathbf{5.15 (\pm 2.75)}$ & $\mathbf{1.86 (\pm 3.06)}$ && $24.03 (\pm 7.50)$ & $\mathbf{13.30 (\pm 5.41)}$ & $\mathbf{6.72 (\pm 3.38)}$ & $\mathbf{0.57 (\pm 1.09)}$\\
        \bottomrule
      \end{tabular}\\ 
    
\end{table*}
To add an attention formalism to our definition of the A2GC-LSTM cell, we adapt the spatial attention formulation for graph convolutions of Si et al. to our problem\cite{si2019attention}. The resulting cell is visualized in Fig.~\ref{cell_figure} with the attention block detailed as the following:
\begin{equation}
    \label{attention_eq}
    \begin{aligned}
        &\mathbf{q}_t       &&= \textit{ReLU}\left(\sum_N\mathbf{H}_{t}\mathbf{W}_a\right)\\
        &\tilde{\mathbf{q}}_t  &&= \tanh(\mathbf{H}_{t}\mathbf{W}_h + \mathbf{q}_{t}\mathbf{W}_q + \mathbf{b}_q)\\
        &\mathbf{\alpha}_t   &&= \sigma(\tilde{\mathbf{q}}_t\mathbf{W}_a + \mathbf{b}_a)\\
        &\tilde{\mathbf{H}}_t   &&= \mathbf{\alpha}_{t}\odot\mathbf{H}_t + \mathbf{H}_t, 
    \end{aligned}
\end{equation}
where the $\mathbf{W}$ are weight matrices, the $\mathbf{b}$ are biases and $\mathbf{H}_t$ is the current hidden state of the A2GC-LSTM. \\
With this $\mathbf{O}_t$ in Equation \ref{lstm_eq} becomes:
\begin{equation}
    \label{lstm_eq_update}
    \begin{aligned}
        &\mathbf{O}_t         &&= \sigma_{out}(\tilde{\mathbf{H}}_t),
    \end{aligned}
\end{equation}
and the hidden state in $\mathbf{X}$ of Equation \ref{A2GC-lstm_eq} becomes $\tilde{\mathbf{H}}_{t-1}$, defining the attention-oriented adjacency adaptive graph convolutional LSTM (A3GC-LSTM).\\

\noindent\textbf{Model architecture and training objective.} 
We define the general layout of our model in Fig.~\ref{model_figure}.
For the setup of our A3GC-IP model we follow the training scheme of Transpose\cite{TransPoseSIGGRAPH2021} by separating the pose learning task into three steps. First, the IMUs are assigned to the corresponding graph nodes and fed into an A3GC network learning the position of the leaf joints. 
These are then concatenated along the feature dimension of each joint with the IMU input and fed to the second network learning the position of all joints. 
After concatenating again with the input data the resulting graph is given as input to a third A3GC network to predict the rotation of all joints, i.e. the $\theta$ parameters of the SMPL model defining the pose. In contrast to Transpose the leaf joints in our model are not head, hands and feet, but the outermost joints of our graph, i.e. head, elbows and knees. 
%While it is counter-intuitive to separate the position learning into two steps when using the graph approach, since the non-leaf joints are also predicted by the first model but just not taken into account when calculating the gradients, we have empirically found this setup to be more accurate for pose estimation.
%\todo{might raise questions without solving ones}
The A3GC networks consist of four layers. The input and output layer is given by A2GCs and the two core layers are bidirectional A3GC-LSTM layers.
We train our model towards global rotations, using an MSE loss function: 
%utilizing longitudinal loss weighting:

\begin{equation}
\label{loss_eq}
    L_\theta = \frac{1}{T}\sum_{t=1}^T\sum_{n=1}^N\sum_{f=1}^F(y_{t,n,f}^{true}-y_{t,n,f}^{pred})^2.
\end{equation}

\noindent where $L_\theta$ is the target loss of SMPL model parameter predictions. The sum over $T$ defines the sequence mean, $N$ denotes the number of joints and $F$ the number of features. For the representation of rotation matrices we chose the conventional 9-dimensional representation, as we empirically found better results for this compared to the reduced 6-dimensional representation of rotation matrices~\cite{Zhou_2019_CVPR} as used by Transpose.\\

\noindent\textbf{Contralateral data augmentation.} A possible bias in captured motion data is a bias originating from a predominant left- or right-handedness of the recorded group of people. 
To lift this bias and at the same time enhance the amount of available training sequences, we can utilize the joint based bilateral symmetry of the human body by applying contralateral data augmentation (CDA) prior to training. 
With CDA we mirror every sequence along the body's main axis with the following procedure. 
We first swap the rotation values of the bilaterally symmetric joints. 
Following the notation in Fig.~\ref{joint_numbers} this means joint 3 is swapped with joint 4 and similarly for the pairs \{(0,1), (8,9), (11,12) and (13,14)\}. Then every rotation itself is mirrored by multiplying the axis angle representation of the rotation with the bilateral rotation mirror vector $[1, -1, -1]$. 
For the synthetic data this mirroring is done before synthesizing the IMU data, and for the real training data, the same scheme is applied to the IMU rotations with the mirrored joint pairs (3,4) and (13,14) bearing IMU sensors. 
For the real IMU data, also the acceleration needs to be mirrored, which is accomplished by multiplying every acceleration vector with the bilateral spatial mirror vector $[-1, 1, 1]$.

\section{Experiments}
We evaluate our method and compare against DIP and Transpose which define the SOTA for sparse IMU-driven pose estimation as well as the methodically comparable approach of G-GRU~\cite{Li_2020_CVPR}. In addition to the comparison against the original DIP, we also compare against a version of DIP using global target rotations instead of local ones, as it is closer to our approach. As our evaluation is focused on the comparison of pose prediction approaches, we exclude the physical inertial poser (PIP)~\cite{PIPCVPR2022} which introduces a non-learnable postprocessing physics module. Since this module could be applied to all compared models, including ours, we decided to not increase the complexity of our evaluation through its inclusion.

To enable a fair comparison of all tested techniques, we have obtained the provided source codes and training scripts, and retrained all techniques. During our retraining, the IMU data is preprocessed following Huang et al.~\cite{huang2018deep}, whereby 5 sensors are normalized with respect to the 6th sensor at the pelvis. We further apply the acceleration scaling by dividing with a factor of $30$ as proposed by Yi et al.~\cite{TransPoseSIGGRAPH2021} as well as the data standardization as used by Huang et al.~\cite{huang2018deep}. For all three methods, we then employ a pre-training phase, during which we train on synthetic IMU data, which we generated from motion capture sequences available through the AMASS motion capture dataset~\cite{AMASS:ICCV:2019}. During this pre-training we exclude all Total Capture sequences, since we lateron test on Total Capture. Afterwards, we finetune by following the original training protocol as well as train/validation split by Huang et al.~\cite{huang2018deep}. Finally, we test on both, DIP-IMU's test set~\cite{huang2018deep} as well as Total Capture~\cite{Trumble:BMVC:2017}. This retraining and evaluation procedure does not only enable a direct comparison, but also allows for constraining the synthetic IMU data generation to only those sequences, not contained in any of the tests sets, a requirement which would otherwise be violated by the original Transpose train/test split, which also contains Total Capture sequences.\\

\noindent\textbf{Other training details.} 
We utilize a standardization of the input and target data based on the training statistics as done by DIP. 
All models are implemented in PyTorch. 
Optimization is done using an Adam optimizer~\cite{kingma2014adam} with an initial learning rate of $0.001$ for training on synthetic IMUs and $0.0001$ for finetuning on real IMU data, and an exponential decay with a rate of $0.8$ applied per epoch. The training is terminated with an early stopping routine after 3 consecutive epochs of no improvement. All models are trained with dropout using a rate of $0.2$ on the input and $0.3$ on the cells hidden state. For our model as well as all compared models operating in the Transpose scheme of separated learning tasks we apply hidden feature dimensions of $256, 64$ and $128$ for the leaf position, full position and pose network respectively and train the three networks separately with a Gaussian noise on the position input of the second and third network with mean $0$ and standard deviations of $0.025$ and $0.04$ respectively.%Other things we could still keep in supplemental were mostly there just because I already had them. Now they would require a lot of retraining, as the old results can no longer be reused. This includes 6D output, adjacency initializations and GRU vs. LSTM.  

\subsection{Quantitative Evaluation}
\begin{table*}[ht]
    \caption{\label{dip_ablation_table}Ablation study on our model. The rows indicate from top to bottom our full model, the same trained without contralateral data augmentation, the model without the attention formalism inside the LSTM cells and the model without adjacency adaptivity inside the LSTM cells.}
    \centering
    \setlength{\tabcolsep}{2pt} % Default value: 6pt
    \renewcommand{\arraystretch}{1} % Default value: 1
    \footnotesize
    \begin{tabular}{l cccc c cccc}
        \toprule
        &\multicolumn{4}{c}{\textbf{DIP-IMU}}
        &&\multicolumn{4}{c}{\textbf{Total Capture}}\\
        \cmidrule{2-5}
        \cmidrule{7-10}
                        & \textbf{DIP Err} $[deg]$
                        & \textbf{Ang Err} $[deg]$
                        & \textbf{Pos Err} $[cm]$
                        & \textbf{Jerk Err} $[\frac{km}{s^3}]$
                        && \textbf{DIP Err} $[deg]$
                        & \textbf{Ang Err} $[deg]$
                        & \textbf{Pos Err} $[cm]$
                        & \textbf{Jerk Err} $[\frac{km}{s^3}]$\\
        %\cmidrule{2-5}
        %\cmidrule{7-10}
        \midrule
          
          A3GC & $13.57 (\pm 6.76)$ & $\mathbf{7.18 (\pm 3.72)}$ & $5.15 (\pm 2.75)$ & $\mathbf{1.86 (\pm 3.06)}$ && $24.03 (\pm 7.50)$ & $\mathbf{13.30 (\pm 5.41)}$ & $\mathbf{6.72 (\pm 3.38)}$ & $\mathbf{0.57 (\pm 1.09)}$\\

          A3GC (-CDA)& $13.99 (\pm 7.39)$ & $7.44 (\pm 4.03)$ & $5.68 (\pm 3.13)$ & $1.90 (\pm 3.11)$ && $24.47 (\pm 8.03)$ & $14.05 (\pm 5.65)$ & $7.50 (\pm 3.71)$ & $0.62 (\pm 1.17)$\\
          
          A3GC (-Attention) & $\mathbf{13.37 (\pm 6.54)}$ & $7.28 (\pm 3.75)$ & $\mathbf{5.11 (\pm 2.62)}$ & $1.89 (\pm 3.12)$ && $\mathbf{23.94 (\pm 7.45)}$ & $13.45 (\pm 5.43)$ & $6.89 (\pm 3.44)$ & $0.68 (\pm 1.31)$\\
          
          A3GC (-Adj. Adapt.) & $14.13 (\pm 6.95)$ & $7.94 (\pm 4.15)$ & $6.61 (\pm 3.34)$ & $1.92 (\pm 3.09)$ && $25.26 (\pm 8.83)$ & $14.02 (\pm 5.98)$ & $8.49 (\pm 4.40)$ & $0.68 (\pm 1.27)$\\
          
        \bottomrule
      \end{tabular}\\ 
\end{table*}
\noindent\textbf{Model accuracy.} We evaluate the methods on four metrics. These are the mean joint angle error with respect to the joints selected for analysis by Huang et al.~\cite{huang2018deep} building on the work of SIP~\cite{von2017sparse}, as well as the mean joint error for angle, position and jerk with respect to all 15 joints. 
The first metric only evaluates the error on the shoulders and hip joints (joints 0, 1, 11 and 12 according to the numbering introduced in Fig.~\ref{joint_numbers}), which is a meaningful decision to analyze the general capability of the model to generalize to the full pose from the IMU measurements, as there is no IMU data directly available for these joints, but they still are part of the extremities and thus a good indicator for the correctness of the pose. 
As this gives no direct information of the total accuracy for the pose estimation, we employ the other three metrics on the complete skeleton. 
Angle and position are used together to give a valid expression of the correctness of the pose. 
The error on the jerk as the third derivative of position, measures the temporal stability of the prediction relative to the ground truth by quantifying effects such as trembling of joints or body parts in the prediction, which are still in the ground truth and vice versa. To get the values for position and jerk, we compute the respective joint positions using the SMPL body model given the true and predicted $\theta$ values. 
The jerk $j$ is obtained from the respective true or predicted positions as a discrete value:

\begin{equation}
    j_t = \frac{p_t-3p_{t-1}+3p_{t-2}-p_{t-3}}{\Delta t^3},
    \label{jerk_eq}
\end{equation}

\noindent where $t$ counts the frames, $p$ is the position and $\Delta t$ is the time per frame.
\noindent We report the comparison of our models with the SOTA in Table~\ref{main_table}, whereby we compare the models after finetuning on the DIP-IMU train and validation split. The errors reported are the mean and standard deviation over all sequences, timesteps and the respective number of joints. 
In addition to the SOTA methods for our problem, we also compare to the method of Li et al. (G-GRU)~\cite{Li_2020_CVPR}, as it is most comparable to our LSTM formulation. As it is build for a different task, we test it by keeping our model architecture and replacing our proposed A3GC-LSTMs with their G-GRU formulation.\\
The first observation we can make is that the proposed A3GC-IP model shows the best scores on almost all metrics. 
The only exclusion from this is the DIP-error on the Total Capture dataset, on which our model is beaten by DIP, while at the same time our model shows the best score on the angle error with respect to all joints by a large margin. 
This is connected to the observation that the overall accuracy by employing global target rotations on DIP is greatly increased, while the accuracy on the Total Capture dataset is decreased. 
From this we can deduct that a model trained towards global target rotations has a significantly better predictive quality on data similar to the training data, but loses some of the generalization capability towards unseen, more different data. 
The fact that the angular error of DIP is very close to the global variant of DIP but the DIP-error being significantly lower, indicates that local target rotations lead to a more even distribution of the error across the skeleton. 
The results of G-GRU show that the simple inclusion of a graph formalism into the network by any method does not increase the accuracy of the model, as it is at best on par with Transpose. 
Thus we can conclude that it is the specific formulation of the A3GC cell leading to a better performance.\\
%\todo{Unsure if we should expand this, as I can hardly do it without bashing on another accepted paper (transpose)}

\noindent\textbf{Number of parameters.}\label{param_comp_section} The formulation of A2GCs has a great benefit compared to the similar graph-based gated recurrent unit (G-GRU) formulation proposed by Li et al.~\cite{Li_2020_CVPR}.
Applied to the data matrices of size $N\cdot F_i$, where $N$ is the number of nodes in the graph and $F_i$ the layers input feature size, we have $(F_i+F_h) \cdot F_o$ parameters in each of the four linear operations of the standard LSTM cell, with the layers hidden state and output feature size $F_h$ and $F_o$. 
By replacing the linear operations with our A2GCs, we increase the number of parameters per operation to $N^2 + (F_i+F_h) \cdot F_o$, thus increasing the parameter count of one complete cell by $4\cdot N^2$. 
The approach of G-GRU applied to an LSTM cell would instead keep the linear operations and add one adjacency adaptive graph convolution on the hidden state, resulting in a total increase of $N^2 + F_h \cdot F_o$. 
Applied to our network we have $F_i=F_h=F_o=F$ for the first hidden LSTM layer and $F_i=F_o=F$ and $F_h=2F$ for the second hidden LSTM layer after concatenation of the forward and backward pass of the former. With our hidden feature size of the three networks given as $256$, $64$ and $128$ respectively we result in a total difference in the number of parameters $\mathcal{N}$ of the two recurrent layers combined for each of the three networks:
\begin{equation}
\mathcal{N}_{\textit{G-GRU}}-\mathcal{N}_{\textit{A2GC-LSTM}}=\begin{cases}390,516 \quad\textit{for}\quad F_h=256\\21,876\quad\textit{for}\quad F_h=64\\95,604\quad\textit{for}\quad F_h=128\end{cases}    
\end{equation}
\noindent In sum for the recurrent layers of all three networks, this difference results in $3,959,436$ parameters for the G-GRU formulation applied to LSTMs and $3,451,440$ parameters for the A2GC-LSTM formulation, a reduction by 14.7\%. 

\subsection{Qualitative Evaluation}
\noindent For the qualitative evaluation we visualized the poses using the SMPL model. As the models predict only the parameters for the $15$ core joints, the remaining $8$ joints for hands and feet are set to identity, while the pelvis joint is fixed. Thus, in all visualizations there is no further rotation applied to these regions.
\begin{figure}[t!]
    \center
    \includegraphics[width=0.95\linewidth]{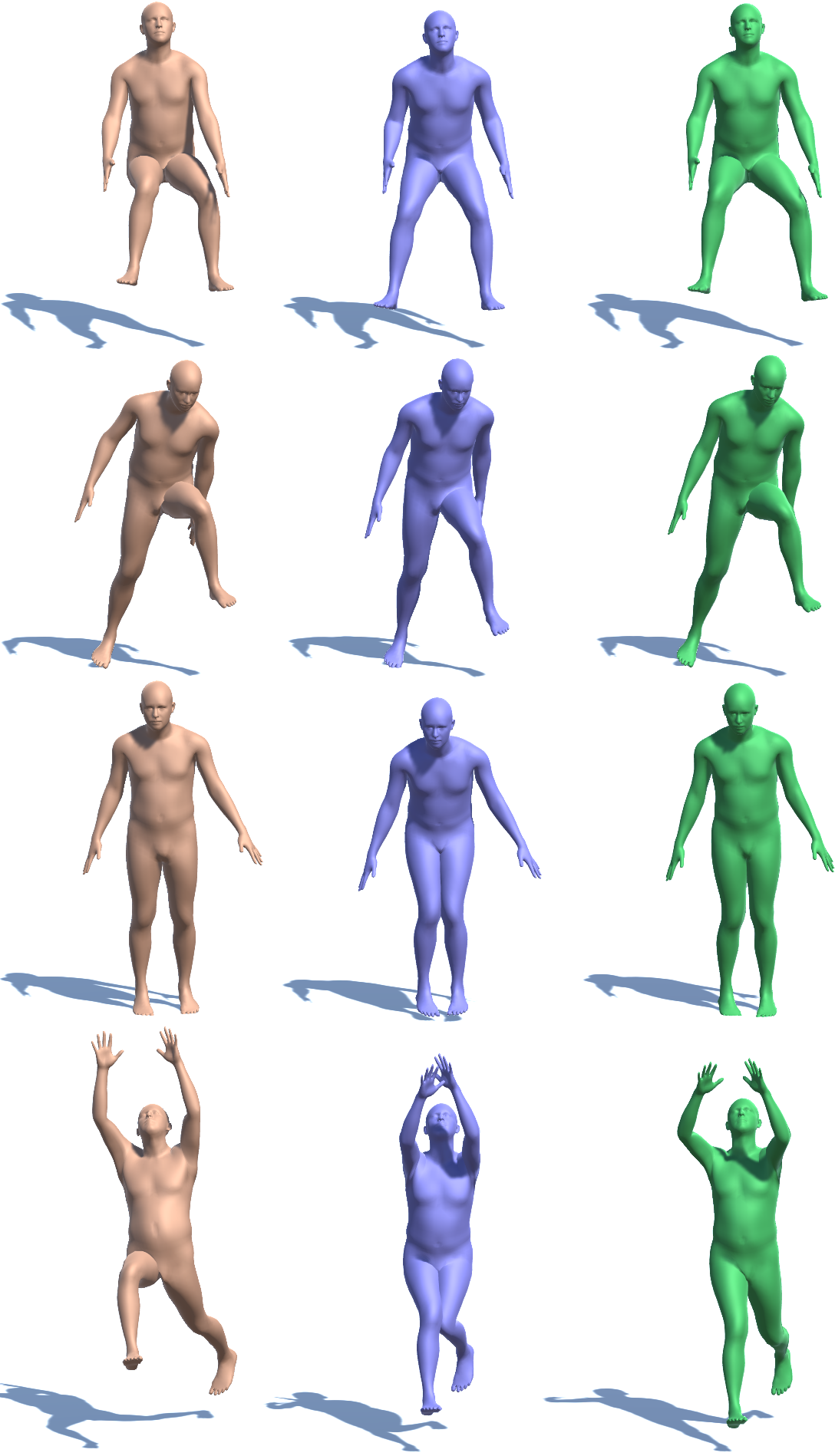}
    \caption*{\raggedright\small \hspace{0.9cm} Ground Truth \hspace{0.95cm} Transpose \hspace{1.75cm} Ours}
    \caption{\label{good_figure} Representative frames among the best scoring poses relative to Transpose from the test datasets. From left to right we show the ground truth, Transpose and our approach for different movements.} 
\end{figure}
In Fig.~\ref{good_figure} we show some examples out of those with the best score of our model relative to Transpose. To select these we sort all frames from best to worst with respect to the relative score between our model and Transpose, from which we selected the $10$ leading examples with the additional constraint that at least $300$ frames are between samples to assert a variation in the shown examples. In the first two rows wee see two example poses with bend legs. Here we observe that our model is much closer to the ground truth than the SOTA, which is most significant in the leg positions. In the third row we show an example of an upright pose where both Transpose and our method introduce a wrong rotation of the knees towards the body center. Nevertheless our model does this to a smaller extent and at the same time manages to keep the unbent pose of the torso, different to the SOTA. In the last row we show an example of a faster movement in form of a jumping pose. While out method shows significant differences to the ground truth here, it is significantly closer to the ground truth than Transpose with respect to the arm, leg and head pose.
\begin{figure}[t!]
    \center
    \includegraphics[width=0.95\linewidth]{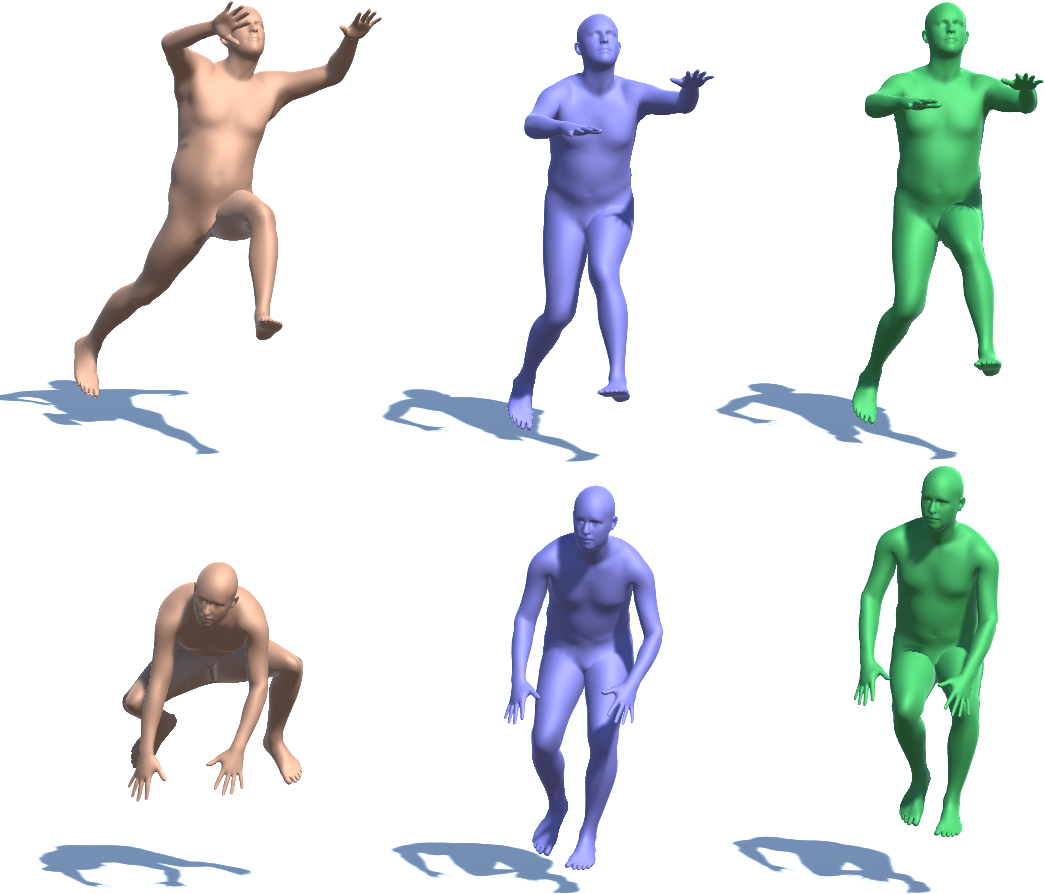}
    \caption*{\raggedright\small \hspace{0.9cm} Ground Truth \hspace{0.95cm} Transpose \hspace{1.75cm} Ours}
    \caption{\label{bad_figure} Representative frames among the worst scoring poses with respect to our model from the test datasets. From left to right we show the ground truth, Transpose and our approach for different movements.} 
\end{figure}
In Fig.~\ref{good_figure} we show two examples out of the worst scoring poses with respect to our model, selected similarly to the previous examples. On top we see an example of a pose in midst of a long jump, which is both rare in the data and very short timed. While our model fails to correctly estimate this pose, it is still visibly closer to the ground truth than Transpose regarding the arm and leg poses. In the second example we observe an example of a rare and rather unnatural movement of crouch-walk using the hands as support on the ground. As movements like this are not part of the DIP-IMU dataset and thus not in the real data used for finetuning, both models fail to predict a pose anywhere close to the ground truth. While our model is a bit closer in the leg pose, at the same time Transpose can predict the forwards bending of teh Torso to a better degree. Thus, visually, we cannot regard any method to be the better one in this case. 
In addition to this short discussion we encourage the reader to watch the supplemental video, in which we show more sequences in animation to get a better understanding of the quality of the predictions, as these are difficult to capture in static images.

\subsection{Ablation Study}\label{Ablation_section}
In this section we analyze the effects of different aspects of our model. 
We detail the results of this ablation study in Table~\ref{dip_ablation_table}. 
The ablations are conducted separately on the contralateral data augmentation (\emph{A3GC -CDA}), the attention formalism introduced inside the LSTM cell (\emph{A3GC -Attention}) and the adjacency adaptivity inside the LSTM cell (\emph{A3GC -Adjacency Adaptivity}), while the rest of the model and training are kept the same. 
We note that for the ablation of adjacency adaptivity the input and output A2GC layers remain unchanged and that this results in a formulation of the LSTM cell similar to Si et al.~\cite{si2019attention}. 
The results show that each of the ablated aspects is vital to the performance of our proposed model.
We further observe that the inclusion of spatial attention inside the LSTM cell leads to a better generalization capability, as the effect on DIP-IMU is rather insignificant, but notable on Total Capture. 
In addition the ablation of adjacency adaptivity inside the LSTM cell shows that the application of spatial attention alone is not sufficient for a good estimation, as the score on all metrics is significantly worse for this cases. Lastly the contralateral data augmentation gives an additional increase in the score on both datasets and all metrics. 

\section{Conclusion and Future Work}
In this paper we have shown that the utilization of the human body graph structure with A3GC-IP leads to better generalization towards pose estimation of unobserved movements. We base this on the following observations. (i) the estimation of the complete skeletal pose showed significant increases in accuracy compared to the prior SOTA. 
This was achieved by (ii) combining the spatio-temporal processing of the sequential movement data in a single step, for which we proposed the A3GC-LSTM cell, which processes both spatial and temporal dependencies of the data in one recurrent cell with a significantly lower amount of parameters needed as compared to existing methods. In addition to this we (iii) report a boost in accuracy by utilizing the bilateral symmetry of the human body through contralateral data augmentation.
All code necessary to reproduce our results, including the trained models is publicly available on GitHub (Link will be provided upon acceptance of the paper, code is supplied to reviewers in supplemental material).
%This benefit of higher accuracy comes at the cost of training time, as the A3GC model is currently significantly slower (factor 30) in training and application due to a missing efficient implementation of the proposed A3GC-LSTM cell, however we note that this is no ultimate bound, as the theoretical complexity of the cell remains comparable to standard LSTM cells and also does not prohibit the application to offline cases.\todo{I was mistaken. No idea were the factor 3 originated in my head. Should we still keep it? Theoretically I don't think it's a problem, but yes, it does sound very bad.}

% or


\appendices

\ifCLASSOPTIONcompsoc
  \section*{Acknowledgments}
\else
  \section*{Acknowledgment}
\fi

The authors would like to thank Manuel Kaufmann and the other DIP authors for providing the SMPL parameters for the Total Capture dataset. This work was supported by the BMG/DLR project AktiSmart-KI with the grant ZMVI1-2520DAT200.

\ifCLASSOPTIONcaptionsoff
  \newpage
\fi



\bibliographystyle{IEEEtran}
\bibliography{mybib}


\begin{IEEEbiography}[{\includegraphics[width=1in,height=1.25in,clip,keepaspectratio]{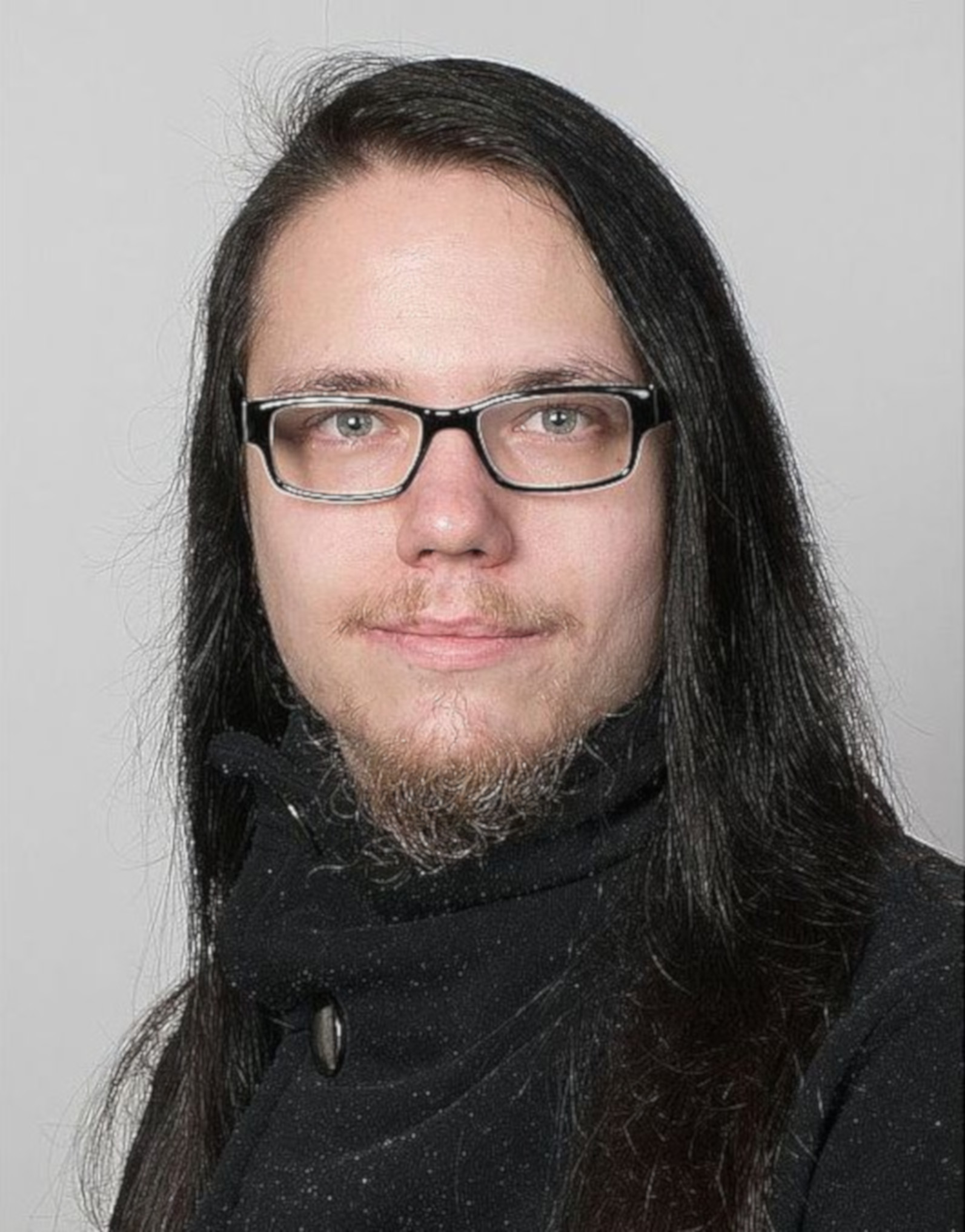}}]{Patrik Puchert}
  received the master's degree in astrophysics from Ludwigs-Maximilians-University Munich in 2018 and is now working as a research associate at the Visual Computing Group at Ulm University.
  His current research interests are in deep learning methods with a focus on human pose estimation and human activity recognition.
\end{IEEEbiography}%\end{IEEEbiographynophoto}

\begin{IEEEbiography}[{\includegraphics[width=1in,height=1.25in,clip,keepaspectratio]{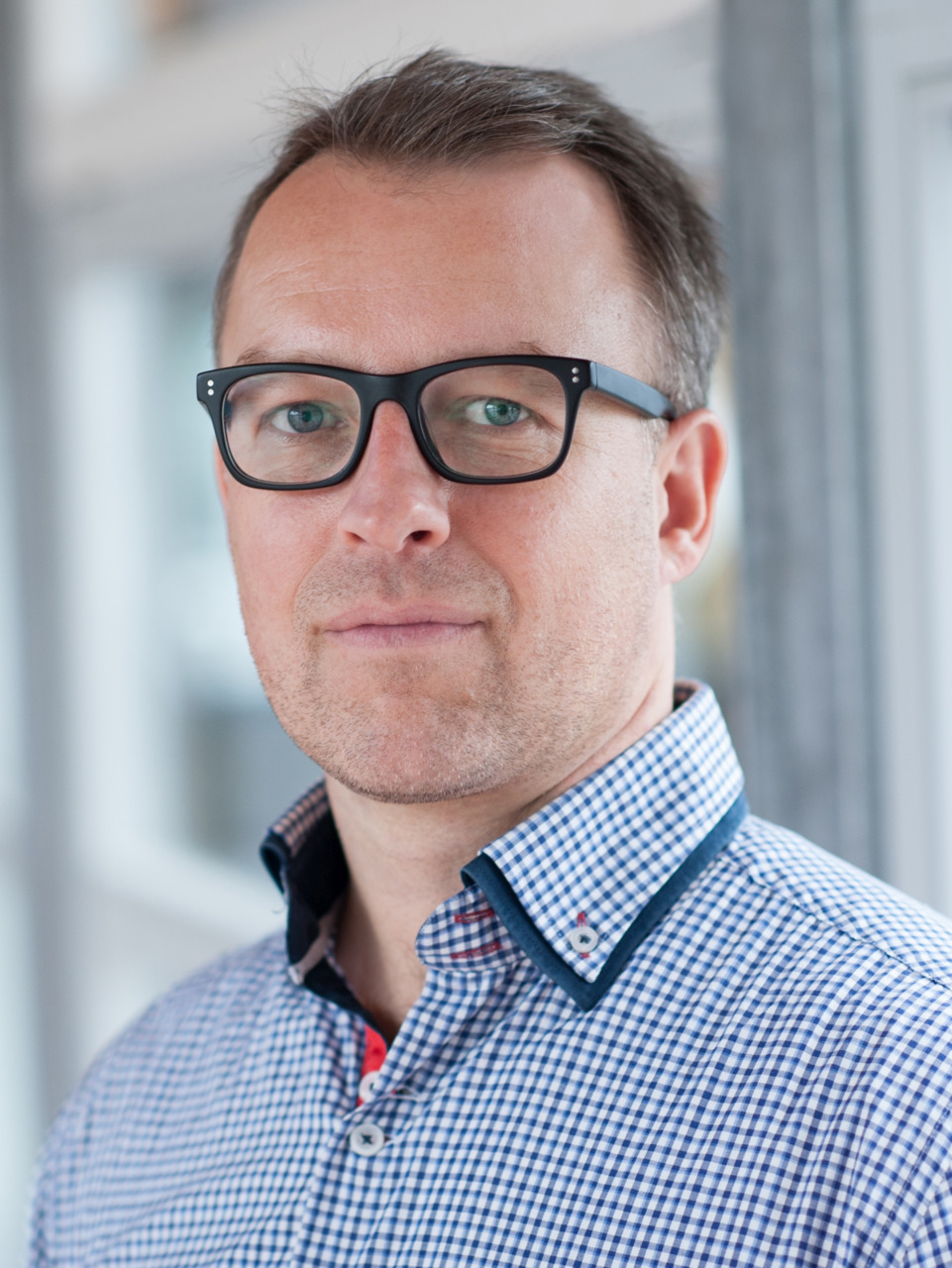}}]{Timo Ropinski}is a professor at Ulm University, where he is heading the Visual Computing Group. Before moving to Ulm he was Professor in Interactive Visualization at Linköping University in Sweden, where he was heading the Scientific Visualization Group. He has received his Ph.D. in computer science in 2004 from the University of Münster, where he has also completed his Habilitation in 2009. Currently Timo serves as chair of the EG VCBM Steering Committee, and as a editorial board member of IEEE TVCG.
\end{IEEEbiography}

%\newpage


%\vfill

\end{document}